\documentclass{article}

\usepackage[preprint]{neurips_2025}

\usepackage[utf8]{inputenc}
\usepackage[T1]{fontenc}

\usepackage{amsmath,amsfonts,amssymb}

\usepackage{graphicx}
\usepackage{booktabs}

\usepackage{hyperref}

\title{FinReflectKG - HalluBench: GraphRAG Hallucination Benchmark for Financial Question Answering Systems}
\author{
  Mahesh Kumar\textsuperscript{1} \quad
  Bhaskarjit Sarmah\textsuperscript{1} \quad
  Stefano Pasquali\textsuperscript{1} \\
  \textsuperscript{1}Domyn Inc \\
  \texttt{\{mahesh.kumar, bhaskarjit.sarmah, stefano.pasquali\}@domyn.com}
}

\begin{document}
\maketitle

\begin{abstract}
As organizations increasingly integrate AI powered question-answering systems into financial information systems for compliance, risk assessment, and decision support, ensuring the factual accuracy and groundedness of AI-generated outputs becomes a critical information systems engineering challenge. Current Knowledge Graph (KG)-augmented QA systems lack systematic methods to detect and handle hallucinations, factually incorrect outputs that can undermine system reliability and user trust. We introduce FinBench-QA-Hallucination, a benchmark for evaluating hallucination detection methods in KG-augmented financial question answering over SEC 10-K filings. Our dataset comprises 755 annotated examples derived from 300 pages, each labeled for groundedness using a conservative evidence-linkage protocol requiring support from both textual chunks and extracted relational triplets. Following an empirical evaluation approach, we systematically assess six detection methods - LLM judges, fine-tuned classifiers, Natural Language Inference (NLI) models, span detectors, and embedding-based approaches - under two controlled conditions: with and without KG triplets. Results show that LLM-based judges and embedding methods achieve the highest performance (F1: 0.82-0.86) in clean conditions, but most approaches exhibit significant degradation when noisy triplets are present, with Matthews Correlation Coefficient (MCC) scores dropping 44-84\% across methods, while embedding-based approaches demonstrated robustness with only 9\% degradation. Statistical analysis via Cochran's Q and McNemar tests confirms significant performance differences (p<0.001). Our findings reveal critical vulnerabilities in current KG-augmented information systems, with direct implications for designing robust financial IS where hallucinations can lead to regulatory violations and flawed decisions. Beyond hallucination detection, our work contributes to IS engineering by demonstrating how AI reliability assessment can be systematically integrated into information system design, providing a template for other high-stakes domains (healthcare, legal, government).
\end{abstract}


\section{Introduction \& Motivation}

Large language models (LLMs) are rapidly being adopted in financial workflows for tasks such as question answering, information extraction, and knowledge graph construction.  In high-stakes domains like finance, however, model outputs that appear fluent and confident can still be factually unsupported by the source documents - a problem commonly referred to as \emph{hallucination}.  Hallucinated answers are especially hazardous in financial settings because downstream decisions (e.g., compliance checks, investment analysis, regulatory reporting) rely on precise numeric values, units, and temporally grounded facts that must be verifiable against primary sources such as SEC 10-K filings.  Despite growing interest in LLM reliability, there remains a lack of focused benchmarks and systematic evaluations that measure how well hallucination detection methods perform when models are supplied with structured KG signals alongside textual context.

This work introduces \textbf{FinBench-QA-Hallucination}, a benchmarking framework and curated dataset designed to evaluate groundedness and hallucination detection in a Knowledge-Graph (KG) assisted question answering pipeline over SEC 10-K source text.  Our dataset comprises 755 annotated Q\&A examples derived from information-rich SEC 10-K pages through an end-to-end pipeline combining automated KG triplet extraction, LLM-based question generation, quality-based page selection to implement negative sampling, and manual validation using a conservative evidence-linkage protocol (detailed methodology in Section~\ref{sec:dataset-construction}).  Each example couples textual context with extracted KG triplets, enabling assessment of how KG retrieval noise-for example, a supporting triplet that contradicts the chunk-affects detection performance across six diverse detection approaches evaluated under controlled conditions (with and without triplets).

There are three core motivational observations that guided our design.  First, KG signals can focus retrieval and reasoning by surfacing concise, structured facts that are likely relevant to a query; when correct and aligned with the source chunk these signals have the potential to reduce uncertainty and improve groundedness detection.  Second, KG pipelines are noisy in practice: extraction errors, schema mismatches, or retrieval misalignments can introduce incorrect triples that appear plausibly relevant to downstream LLMs.  A robust detector therefore must not only recognize unsupported model claims in plain text, but also be resilient to or aware of noisy KG evidence.  Third, human annotation for groundedness is expensive and domain expertise is unevenly available; designing an annotation protocol that privileges explicit evidence linkage (triplet + supporting lines) enables scalable validation while making the groundedness decision less reliant on deep subject-matter expertise.

Building on these observations, FinBench-QA-Hallucination is intended as a focused, reproducible testbed to \textbf{benchmark hallucination detection approaches} and assess their \textbf{robustness to KG signal quality}. Specifically, we address two research questions: (1) how well do existing detection strategies distinguish grounded answers from hallucinations when both textual chunks and KG signals are provided, and (2) how sensitive are those detectors to retrieval noise introduced by incorrect or contradictory triplets? To address these questions we evaluate a diverse set of detection approaches-including LLM-as-judge methods, fine-tuned classifiers, NLI models, span detectors, and embedding-similarity based approaches-under two controlled evidence conditions: \emph{With Triplets} and \emph{Without Triplets} (text-only).

While the FinReflectKG pipeline includes automated validation and correction mechanisms to identify and fix problematic triplets, we applied these loosely rather than strictly, and did not perform human verification of extracted triplets. This design choice enables assessment of detector robustness to naturally occurring KG extraction errors without synthetic noise injection. Strict triplet verification would limit error diversity and reduce ecological validity; our approach reflects realistic deployment conditions where KG pipelines produce imperfect but plausible structured evidence.

Our contributions are as follows:
\begin{itemize}
  \item We introduce FinBench-QA-Hallucination, a curated benchmark of 755 annotated Q\&A examples derived from SEC 10-K pages that explicitly couples generated answers with model-reported supporting triplets and supporting textual lines.
  \item We define a conservative, evidence-linkage based annotation protocol and apply it to produce binary groundedness labels together with categorical rejection flags that isolate common failure modes (bad triplet, insufficient evidence, unit errors, ambiguous questions).
  \item We provide a controlled experimental framework that isolates the effect of KG signals and retrieval noise by comparing detector performance in \emph{With Triplets} (matched vs. contradicted) and \emph{Without Triplets} settings across a suite of detection methods.
  \item We analyze practical limitations and failure modes of KG-assisted detection - including the influence of noisy triples and the annotator expertise tradeoffs - and discuss implications for deploying grounded QA systems in finance.
\end{itemize}

By making the dataset design, annotation protocol, and evaluation conditions explicit and reproducible, this paper aims to advance rigorous, bias-aware assessment of hallucination detectors in KG-augmented financial QA systems and to inform safer, more transparent deployment practices for LLMs operating over regulatory filings.

\section{Related Work}
Hallucination i.e. the generation of plausible but factually incorrect or unverifiable content, poses a critical challenge for deploying LLMs in high-stakes applications. Research has focused on benchmarks, detection methods, and mitigation strategies. The intersection of hallucination detection, knowledge graph-augmented retrieval, and financial AI represents an emerging frontier. This section reviews work across three areas: (1) hallucination detection in LLMs, (2) KG-augmented QA systems, and (3) financial NLP applications, concluding with the research gap motivating our benchmark.

Several benchmarks evaluate LLM factuality. \cite{li2024dawn} introduced HaluEval 2.0 with an empirical taxonomy showing hallucinations arise from training biases, architecture limitations, and sampling strategies. \cite{li2023helma} developed HELMA with human-annotated examples via sampling-then-filtering, demonstrating external knowledge integration and multi-step reasoning improve detection accuracy.

For fine-grained evaluation, \cite{hu2024refchecker} proposed RefChecker, decomposing outputs into claim-triplets verified against reference documents. On an 11,000-claim benchmark, claim-level checking substantially outperformed sentence/document-level approaches. \cite{chen2023factchd} introduced FactCHD covering diverse factuality patterns and multi-hop evidence chains, emphasizing explanation-aware detection for complex reasoning. \cite{gu2024anah} developed ANAH-v2, an EM-style self-training pipeline producing automated annotators outperforming GPT-4 on HaluEval and HalluQA, demonstrating scalable methods reduce human effort while maintaining quality.

Detection methods span three families: self-consistency, retrieval-based, and internal-state analysis. Self-consistency methods exploit that hallucinations exhibit lower consistency across generations. \cite{manakul2023selfcheckgpt} proposed SelfCheckGPT, generating multiple responses and measuring consistency via BERTScore, QA, and n-gram overlap, though requiring multiple costly inference passes. Retrieval-based methods verify outputs against external sources; RefChecker extracts and verifies claims against retrieved documents. Internal-state methods analyze activations without external references. \cite{su2024unsupervised} proposed MIND, monitoring internal states to identify uncertainty patterns, showing hidden dynamics contain factuality signals. \cite{mir2025geometry} introduced LSD (Layer-wise Semantic Dynamics), tracking semantic consistency across layers to detect "semantic drift" where representations diverge from factual content, achieving state-of-the-art performance.

Recent work explores specialized domains. \cite{pandit2025medhallu} developed MedHallu for medical hallucinations, showing general-purpose detectors fail due to terminology complexity and reasoning requirements, motivating domain-specific benchmarks. However, existing benchmarks primarily target open-domain or medical applications, not KG-augmented financial QA where structured triplets must integrate with detection mechanisms while satisfying domain requirements (numerical accuracy, temporal validity, regulatory compliance).

Knowledge graphs provide structured entity-relationship representations complementing unstructured text retrieval. Recent work integrates KGs with LLMs for improved factuality and reasoning. \cite{mavromatis2024gnn} introduced GNN-RAG, encoding KG subgraphs via GNNs and conditioning generation on textual and graph-structured evidence, improving reasoning accuracy over text-only retrieval by 8-12\% on CommonsenseQA and WebQSP. \cite{he2024g} proposed G-Retriever, employing graph-aware retrieval to identify relevant subgraphs by question semantics then linearizing structure for LLM consumption, achieving strong performance while maintaining interpretability. \cite{ma2024think} developed Think-on-Graph 2.0, iteratively refining KG retrieval via beam search-inspired exploration of reasoning paths, reducing hallucinations by grounding steps in explicit graph edges. These KG-RAG systems demonstrate structured knowledge integration potential but focus on accuracy metrics for standard benchmarks, not hallucination detection performance or robustness to KG noise. Our benchmark provides controlled experiments with clean vs. noisy triplets.

The financial domain presents unique NLP challenges: specialized terminology, numerical reasoning, temporal dynamics, and high-stakes contexts. \cite{araci2019finbert} introduced FinBERT, pre-trained on financial news and SEC filings, improving sentiment analysis and NER. \cite{wu2023bloomberggpt} developed BloombergGPT, a 50B parameter model achieving state-of-the-art financial NLP performance while maintaining general capabilities. \cite{yang2023investlm} proposed InvestLM using financial domain instruction tuning, showing substantial improvements on reasoning tasks (portfolio optimization, risk assessment, market analysis) over general-purpose LLMs.

Financial benchmarks evaluate LLM capabilities. \cite{NEURIPS2023_6a386d70} introduced Pixiu with 136,000 samples across nine tasks (sentiment analysis, NER, QA, numerical reasoning), providing broad coverage. \cite{xie2024finben} proposed FinBen targeting long-form SEC analysis with multi-hop questions requiring cross-document synthesis, revealing state-of-the-art LLMs achieve only 60-70\% accuracy on complex financial reasoning.

Despite progress across these areas, a critical gap remains: no work systematically evaluates hallucination detection in KG-augmented financial QA. Hallucination benchmarks (HaluEval, HELMA, RefChecker, FactCHD) are domain-general or medical-focused, using textual references without incorporating KG evidence or examining KG-specific failures (entity linking errors, relation noise, temporal misalignment). KG-augmented systems (GNN-RAG, G-Retriever, Think-on-Graph 2.0) improve reasoning but focus on accuracy metrics for standard benchmarks (CommonsenseQA, WebQSP), not hallucination detection or KG quality effects. Financial benchmarks (FinBen, Pixiu) provide SEC-derived tasks but rarely couple KG retrieval with hallucination detection.

Financial QA requires high provenance, numerical fidelity, and regulatory compliance. KG augmentation offers verifiable structured evidence but introduces failure modes (entity linking errors, relation noise, temporal misalignment). Existing detectors (RefChecker, MIND, LSD) provide promising techniques but lack evaluation in financial KG-RAG contexts. Understanding detector performance with noisy KG evidence is essential for trustworthy financial AI. Our work introduces FinBench-QA-Hallucination to evaluate hallucination detection in KG-augmented financial QA.

\section{System Design}

Our pipeline consisted of the following, with emphasis on the parts relevant to dataset creation and evaluation:

\begin{itemize}
  \item \textbf{Source documents:} SEC 10-K (and related) filings were the corpus used to generate the dataset. From these filings we produced text \emph{chunks} and extracted relational \emph{triplets}.
  \item \textbf{Chunking and Triplet extraction:} We leverage the FinReflectKG methodology~\cite{finreflectkg}, which employs structure-aware chunking tailored to financial filings followed by LLM-based (Qwen-235B) extraction. We adopted the reflection-driven extraction mode, which uses an iterative feedback-correction loop with critic and correction LLMs to generate triplets, followed by automatic schema validation. The extracted triplets were not manually verified; our focus was on creating a hallucination benchmark dataset from financial texts to evaluate detection approaches, with manual verification applied only to the generated Q\&A pairs (see Section~\ref{sec:manual-verification}). The framework extracts structured subject-relation-object triples using 24 predefined entity types (ORG, FIN\_METRIC, SEGMENT, STATEMENT, etc.) and 29 relationship types (Guides\_On, Has\_Value\_Merged, Part\_Of, Belongs\_To\_Segment, etc.) from a financial domain vocabulary. We extended the schema to encode numeric values with their units and temporal context in a MERGED\_NUMERIC\_UNIT\_PERIOD format (e.g., \texttt{["PG", "ORG", "Guides\_On", "4\%-FY-2024", "MERGED\_NUMERIC\_UNIT\_PERIOD"]}). Granular extraction methodology details are out of scope; the high-level dataset construction process is described in Section~\ref{sec:dataset-construction}.
  \item \textbf{Knowledge graph (FinReflectKG):} Triplets were collected and intended for ingestion into a knowledge graph referred to as FinReflectKG. No details about graph schema, storage backend, or ingestion pipeline were provided; therefore we only note that the triplets served as KG signals during generation and later evaluation.
  \item \textbf{Question generation model:} Qwen-3-235B was used to generate (Question, Answer, Reasoning) tuples. For each sampled page, 5--7 triples were randomly sampled (when available) and provided to the model together with the page chunk; the model returned one or more Q\&A examples, the supporting triplet, and one or more supporting lines (textual snippets) from the chunk that the model asserted supported the answer.
  \item \textbf{Manual validation app:} An internal web application was used by human validators to inspect generated examples. Validators reviewed the record containing (Context chunk, Sampled Triples, Question, Answer, Supporting Triplet, Supporting Lines, Model Reasoning) and applied an annotation protocol (see Section~\ref{sec:manual-verification}).
\end{itemize}

\section{Hallucination Detection Framework}
\subsection{Definitions and Taxonomy}
\label{sec:definitions}

\begin{itemize}
  \item \textbf{Groundedness:} an answer is considered \texttt{Grounded} only if it is correct with respect to the source chunk and is supported by both (a) the supporting triplet identified by the generator, and (b) the supporting lines (textual snippet(s)) from the chunk.
  \item \textbf{Hallucination:} an answer is considered a hallucination if it is unsupported by the provided context (the chunk and supporting lines) or if the supporting triplet contains incorrect values such that the answer cannot be established from the supplied evidence.
\end{itemize}

\subsection{Experimental Design: Robustness to KG Signal Quality}
\label{sec:triplet-robustness}

To assess detector robustness to KG extraction errors, we evaluated all 755 examples under two conditions: \textbf{Without Triplets} (text-only baseline) and \textbf{With Triplets} (text + KG triplets from FinReflectKG). This controlled experiment leverages naturally occurring extraction errors without synthetic noise injection. For cases where predictions differed between conditions, we categorized outcomes as \textbf{Helped} (triplet enabled correct prediction) or \textbf{Hurt} (triplet caused failure), with Net Impact (Helped - Hurt) quantifying robustness to real KG noise. Quantitative results and analysis are presented in Section~\ref{sec:results}.

\subsection{Detection Methods}
The following detection methods were evaluated.

\begin{enumerate}
  \item \textbf{LLM as Judge:} The pre-trained Qwen-3-235B model~\cite{qwen235b} was used as a scoring judge outputting 1--5 scale. Input consisted of (Context + Question + Answer + supporting evidence when available). Output score is normalized to [0, 1] for thresholding. To validate findings independently of the generation model, GPT-OSS-120B~\cite{gptoss120b} was evaluated using the identical prompt and scoring protocol.

  \item \textbf{Fine-Tuned LLM as Judge:} Finetuned Lynx-8B~\cite{lynx8b} was used as a classifier on labeled examples to predict binary PASS/FAIL for \texttt{Grounded} vs \texttt{Hallucinated}.

  \item \textbf{NLI Model:} A pre-trained NLI model~\cite{debertav3nli} (DeBERTa-v3) was applied using premise-hypothesis pairs (Premise: context text, Hypothesis: question + answer) producing probabilities for three classes: entailment, neutral, and contradiction. For binary classification, we use \textbf{entailment} threshold (higher score = grounded) and \textbf{contradiction} threshold (lower score = grounded), converting 3-class probabilities to binary decisions.

  \item \textbf{Fine-Tuned Classifier:} A fine tuned hallucination detection classifier~\cite{vectara} trained on (Context, Question, Answer) feature representations; its numeric outputs (factual consistency scores in [0, 1]) were thresholded to obtain binary labels.

  \item \textbf{Span-Based Classifier:} A span detector (LettuceDetect~\cite{lettucedetect}) tries to find hallucinated answer spans (with respect to question and context). Per-span confidence scores are aggregated using \textbf{min} (most conservative), \textbf{median} (balanced), and \textbf{max} (most optimistic) to produce example-level predictions, which are then thresholded for binary decisions.

  \item \textbf{Embedding Similarity:} Per-line context embeddings were computed and compared with the answer embedding using Qwen-0.6B~\cite{qwen06b}. Per-sentence cosine similarities between context lines and the answer embedding are aggregated using \textbf{min}, \textbf{median}, and \textbf{max} to obtain example-level scores. Higher aggregated similarity indicates better grounding; these scores are then thresholded for binary labels. For independent validation, Stella-400M~\cite{stella400m} was evaluated using the same aggregation protocol.
\end{enumerate}

These approaches provide diverse coverage of hallucination detection paradigms. LLM-as-judge methods (Qwen, GPT-OSS, Lynx) represent emerging autonomous evaluation techniques, with Lynx fine-tuned for financial tasks. NLI models offer generic entailment and contradiction assessment without domain-specific training. Vectara and LettuceDetect represent dedicated hallucination detection using distinct mechanisms: classification-based consistency scoring versus span-level detection. Embedding-based similarity methods (Qwen, Stella) are well-suited for short-answer grounding where responses map to specific context segments. This methodological diversity enables comprehensive assessment across different architectural paradigms and inference strategies.

\subsection{Implementation Details}
\label{sec:implementation}

All experiments used PyTorch 2.9.0 and Transformers 4.57.1. Table~\ref{tab:infrastructure} specifies hardware, precision, and modified hyperparameters for each detection method. All other parameters use default values from model configurations.

\begin{table}[ht]
\centering
\caption{Infrastructure and Hyperparameter Specifications for Detection Methods}
\label{tab:infrastructure}
\small
\begin{tabular}{l l l l}
\toprule
\textbf{Method} & \textbf{Hardware} & \textbf{Precision} & \textbf{Modified Parameters} \\
\midrule
Qwen LLM judge & vLLM endpoint & float16 & temperature=0.1 \\
GPT-OSS LLM judge & vLLM endpoint & float16 & temperature=0.1 \\
Lynx LLM judge & NVIDIA T4 & float16 & max\_new\_tokens=600, use\_cache=False \\
NLI DeBERTa-v3 & Apple M4 Pro & float16 & -- \\
Vectara classifier & Apple M4 Pro & float32 & trust\_remote\_code=True \\
LettuceDetect & Apple M4 Pro & float32 & device=mps \\
Qwen Embeddings & Apple M4 Pro & bfloat16 & padding\_side=left \\
Stella Embeddings & Apple M4 Pro & float32 & padding\_side=left \\
\bottomrule
\end{tabular}
\end{table}

\section{Dataset Construction and Annotation}
\label{sec:dataset-construction}

\subsection{Data generation process}

Our dataset construction followed a multi-stage pipeline starting from S\&P 100 companies' fiscal year 2024 SEC 10-K filings (see Appendix~\ref{sec:company-coverage} for the complete list of 57 companies in the final dataset). First, we extracted knowledge graph triplets from all pages in these filings using the FinReflectKG pipeline with Qwen-235B (see Section 3 for extraction framework details). From the resulting corpus of thousands of pages, we filtered to retain only pages containing at least 10 extracted triplets, ensuring sufficient informational density for meaningful question generation. We then iteratively sampled unique pages from this filtered set (companies could repeat but pages could not) and for each sampled page randomly selected 5--7 triplets to provide to the generation model, continuing until approximately 5{,}000 Q\&A pairs were produced. Next, we applied an LLM-as-judge quality filter (details in Section~\ref{sec:manual-verification}) to identify problematic examples. To implement negative sampling and increase representation of challenging cases, we selected 300 pages where at least one Q\&A pair received a \texttt{Drop} flag from the judge; these 300 pages contained 1{,}302 Q\&A pairs in total. Finally, manual validation by human annotators and subsequent filtering for complete detector outputs yielded 755 validated examples (Table~\ref{tab:dataset_stats}), representing the intersection where (1) human annotation is available and (2) all six detection approaches produced valid outputs in both experimental conditions. No document section filtering was applied; all 10-K content (Risk Factors, MD\&A, Financial Statements, etc.) was included to capture diverse content complexity.

\begin{enumerate}
  \item \textbf{Page filtering and iterative sampling:} Pages with fewer than 10 triplets were excluded to ensure informational density. Remaining pages were sampled iteratively (unique pages, companies could repeat) with 5--7 triplets randomly selected per page for question generation.
  \item \textbf{Triple sampling and prompting:} For each sampled page, 5--7 triples were randomly sampled to balance generation diversity with annotation feasibility while enabling challenging cross-contamination scenarios. The page chunk and sampled triples were provided to Qwen-3-235B (temperature=0.1; see Appendix~\ref{sec:qa-gen-prompt}), which generated (Question, Answer, Reasoning) examples and indicated:
    \begin{itemize}
      \item the \emph{supporting triplet} (one triplet from the provided 5--7) it relied on, and
      \item one or more \emph{supporting lines} (textual snippet(s) from the chunk) that the model asserted justified the answer.
    \end{itemize}
    The supporting triplet and supporting lines were required (by the generation protocol) to be tightly coupled to the generated Q\&A and the model's reasoning. Answerability from the knowledge graph was ensured through explicit prompt constraints (detailed in Appendix~\ref{sec:qa-gen-prompt}) and subsequent LLM-as-judge filtering (described in Section~5.1.1).
\end{enumerate}

\subsubsection{Pre-filtering and page selection}
We applied automated LLM-as-judge pre-filtering (Qwen-235B; see Appendix~\ref{sec:qa-eval-prompt}) to assign binary \texttt{Keep}/\texttt{Drop} labels based on question quality (genericity, missing context), answer quality (unsupported claims, assumptions), and overall factuality. The prompt instructed ``When in doubt, DROP'' to ensure high recall of problematic examples. We selected pages where at least one Q\&A received a \texttt{Drop} label, implementing negative sampling to increase representation of challenging cases. All automated labels were hidden from human annotators to ensure independent groundedness decisions.

\subsection{Manual verification and annotation}
\label{sec:manual-verification}
Annotation was performed in an internal web application (Figure~\ref{fig:annotation_interface} in Appendix) by 9 human validators, all in-house AI researchers and engineers with some finance experience. Each example was annotated by a single validator following the conservative evidence-linkage protocol (Section~\ref{sec:definitions}) and rejection rules detailed below, recording binary groundedness labels (\texttt{Grounded}/\texttt{Hallucinated}), optional free-text reasoning, and categorical issue flags.

\paragraph{Rejection rules used by annotators:}
The following rejection rules were explicitly provided and applied during manual verification. Rejected questions were filtered from analysis because usually they were too simple or too ambiguous but Answer rejection means that Answer is not Grounded or is hallucinated.

\begin{itemize}
  \item \textbf{Reject Question:} Question is too generic/ambiguous and can be applied to any company/year/metric without specificity.
  \item \textbf{Reject Answer:} Answer does not include required units when context explicitly specifies values in bn/mn/Billion/Million/\% etc.
  \item \textbf{Reject Answer:} Answer assumes units (million/billion/\%/USD etc.) when context does not explicitly state the unit.
  \item \textbf{Reject Answer:} Answer is not sourced from provided context \textbf{OR} the supporting triplet contains incorrect values.
  \item \textbf{Reject Answer:} Answer might be correct but it cannot be determined from the given context (i.e., it is not grounded in the context - insufficient information available).
\end{itemize}

\subsection{Dataset Composition}
Table~\ref{tab:dataset_stats} presents dataset statistics. 42\% of generated examples were filtered, primarily due to missing human annotations (36\%) and rejected questions (5\%). The valid dataset of 755 examples contains 513 grounded (68\%) and 242 hallucinated (32\%) Q\&A pairs.

\subsection{Data Availability}
The FinBench-QA-Hallucination dataset, including all 755 annotated question-answer pairs, evaluation scripts, and detector implementations, will be made publicly available on GitHub upon publication. This includes raw benchmark results, ground truth annotations, and code to reproduce all reported metrics.

\begin{table}[ht]
\centering
\label{tab:dataset_stats}
\small
\begin{tabular}{l r r}
\toprule
\textbf{Category} & \textbf{Count} & \textbf{\%} \\
\midrule
\multicolumn{3}{l}{\textbf{Dataset Composition:}} \\
Total Generated & 1,302 & 100.0 \\
\quad Valid & 755 & 58.0 \\
\quad Filtered & 547 & 42.0 \\
\midrule
\multicolumn{3}{l}{\textbf{Valid Samples by Label:}} \\
Grounded (Correct) & 513 & 68.0 \\
Hallucinated (Incorrect) & 242 & 32.0 \\
\midrule
\multicolumn{3}{l}{\textbf{Filtering Breakdown (547 filtered):}} \\
\quad Missing human review & 471 & 36.2 \\
\quad Rejected questions & 59 & 4.5 \\
\quad Approach errors & 12 & 0.9 \\
\quad Invalid answer labels & 3 & 0.2 \\
\midrule
\multicolumn{3}{l}{\textbf{Company Coverage (by ticker):}} \\
\quad Unique tickers (1,302 total) & 67 & -- \\
\quad Unique tickers (755 valid) & 57 & -- \\
\bottomrule
\end{tabular}
\newline

\caption{Dataset Statistics (n=755). 755 validated pairs (68\% grounded, 32\% hallucinated) from 57 S\&P 100 companies, annotated by 9 reviewers. Filtered: 471 missing reviews, 59 rejected questions.}
\end{table}

\section{Results \& Analysis}
\label{sec:results}

\subsection{Detection Performance}
All 755 examples were evaluated under the two conditions described in Section~\ref{sec:triplet-robustness} (\textbf{With/Without Triplets}). We report best performance across thresholds (0.05-0.95, 0.05 increments) for F1, MCC, ROC-AUC, and PR-AUC at individually optimized thresholds. We note that for the robustness analysis, we retained the optimal thresholds from the 'Without Triplets' condition, as individually optimizing for the 'With Triplets' condition yielded negligible impact (max $\Delta$F1 < 0.003). Statistical significance assessed using Cochran's Q (omnibus) and McNemar's test (pairwise, Holm-Bonferroni correction), with bootstrap CIs (10,000 iterations, stratified).

Table~\ref{tab:best_scores} presents best detection performance. For each approach, we report F1, MCC, ROC-AUC, and PR-AUC at their individually optimized thresholds in both conditions (W/O = without triplets, W/ = with triplets). Precision and Recall discussed in text at best F1 thresholds.

\begin{table}[ht]
\centering
\caption{Best Detection Performance (n=755). W/O = without triplets, W/ = with triplets. Each metric at individually optimized threshold. \underline{Underlined} = best W/O, \textbf{Bold} = best W/.}
\label{tab:best_scores}
\footnotesize
\begin{tabular}{l l | r r | r r | r r | r r}
\toprule
\textbf{Approach} & \textbf{Measure} & \multicolumn{2}{c|}{\textbf{F1}} & \multicolumn{2}{c|}{\textbf{MCC}} & \multicolumn{2}{c|}{\textbf{ROC-AUC}} & \multicolumn{2}{c}{\textbf{PR-AUC}} \\
& & W/O & W/ & W/O & W/ & W/O & W/ & W/O & W/ \\
\midrule
lynx\_llm\_judge & NA & 0.831 & 0.809 & 0.374 & 0.185 & 0.657 & 0.556 & 0.756 & 0.705 \\
qwen\_llm\_scoring\_judge & NA & \underline{0.863} & 0.818 & \underline{0.505} & 0.205 & \underline{0.722} & 0.550 & \underline{0.794} & 0.702 \\
gpt\_oss\_llm\_judge & NA & 0.849 & 0.806 & 0.493 & 0.159 & 0.718 & 0.547 & 0.791 & 0.699 \\
nli\_deberta3 & contradiction & 0.809 & 0.808 & 0.177 & 0.117 & 0.565 & 0.537 & 0.709 & 0.696 \\
vectara\_hallucination & NA & 0.848 & 0.810 & 0.435 & 0.242 & 0.675 & 0.613 & 0.770 & 0.736 \\
lettuce\_detect & min & 0.827 & 0.809 & 0.343 & 0.054 & 0.662 & 0.513 & 0.760 & 0.685 \\
line\_embeddings (Qwen) & max & 0.824 & \textbf{0.824} & 0.317 & \textbf{0.288} & 0.650 & \textbf{0.644} & 0.755 & \textbf{0.751} \\
stella\_embeddings & max & 0.820 & 0.809 & 0.304 & 0.264 & 0.640 & 0.619 & 0.748 & 0.737 \\
\bottomrule
\end{tabular}
\end{table}

\begin{figure}
    \centering
    \includegraphics[width=1\linewidth]{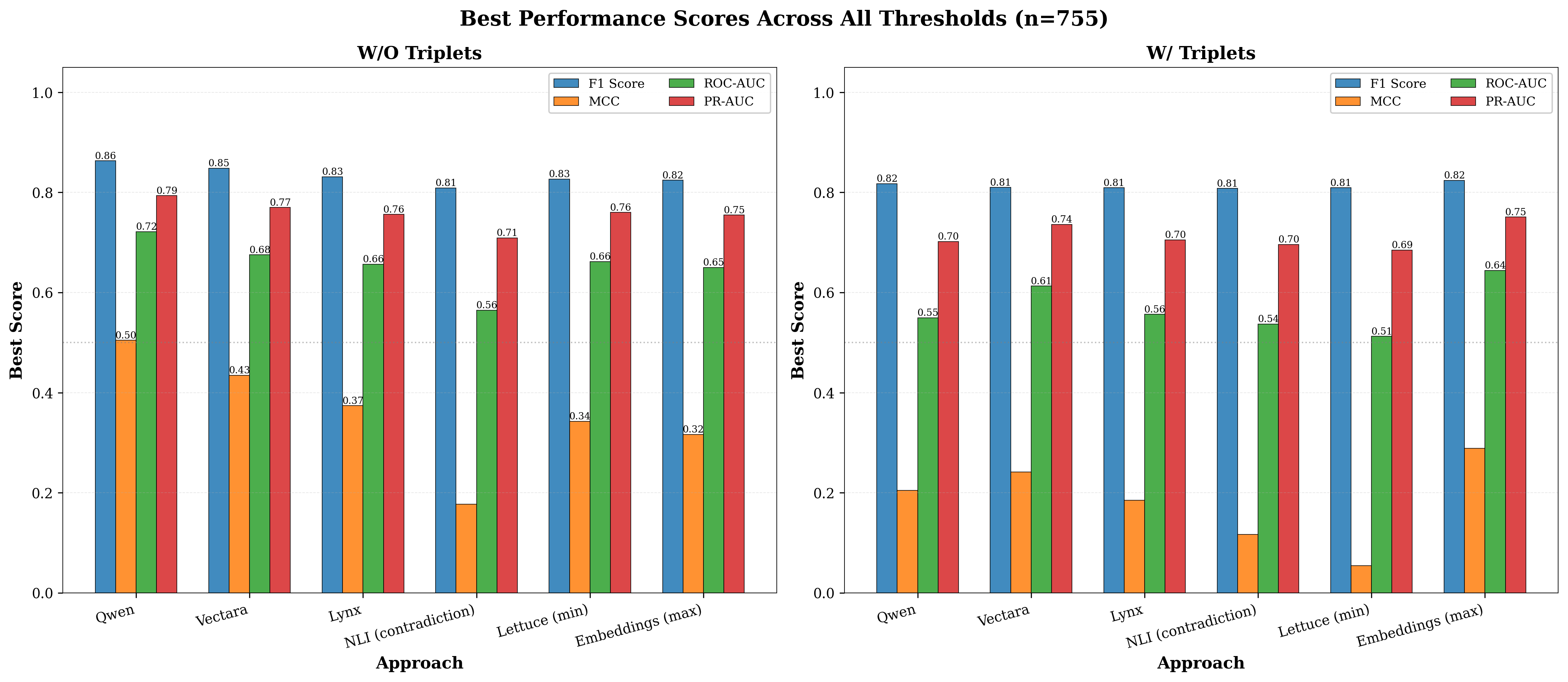}
    \caption{Performance across methods}
    \label{fig:best_scores_bar_chart}
\end{figure}

Table~\ref{tab:best_scores} reveals that most detectors exhibit significant performance degradation when triplets are present, with Qwen's F1 dropping from 0.863 to 0.818 and MCC from 0.505 to 0.205. To understand this degradation pattern, we analyze detector robustness to individual triplet errors below.

\begin{figure}
    \centering
    \includegraphics[width=0.8\linewidth]{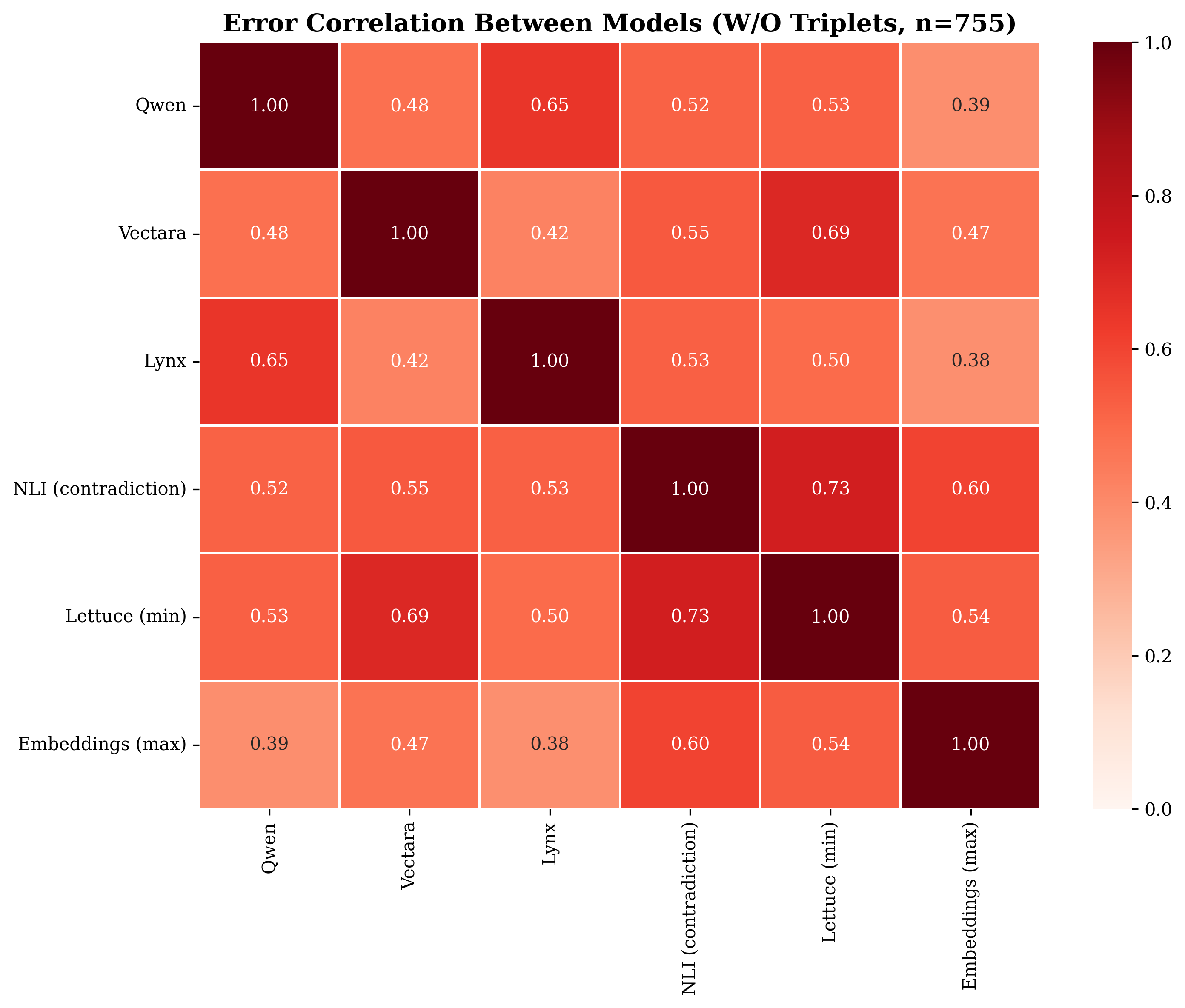}
    \caption{All Methods Error Correlation}
    \label{fig:error_correlation_matrix}
\end{figure}

\subsection{Robustness Analysis}

Table~\ref{tab:triplet_effect} presents the robustness profile for each detector, revealing that \textbf{most detection methods lack robustness to noisy KG signals}, with triplets causing more harm than benefit.

\begin{table}[ht]
\centering
\caption{Detector Robustness to Real KG Extraction Errors (n=755). ``Helped'' = triplet enabled correct prediction. ``Hurt'' = triplet caused failure. Net Impact = Helped - Hurt.}
\label{tab:triplet_effect}
\small
\begin{tabular}{l r r r r}
\toprule
\textbf{Detector} & \textbf{Disagreements} & \textbf{Helped} & \textbf{Hurt} & \textbf{Net Impact} \\
\midrule
Qwen LLM judge & 113 (15.0\%) & 20 & 93 (12.3\%) & -73 \\
Lynx LLM judge & 113 (15.0\%) & 37 & 76 (10.1\%) & -39 \\
NLI DeBERTa (contradiction) & 15 (2.0\%) & 6 & 9 (1.2\%) & -3 \\
Vectara & 114 (15.1\%) & 24 & 90 (11.9\%) & -66 \\
LettuceDetect (min) & 67 (8.9\%) & 15 & 52 (6.9\%) & -37 \\
Line Embeddings (max) & 53 (7.0\%) & 22 & 31 (4.1\%) & -9 \\
\bottomrule
\end{tabular}
\end{table}

Qwen achieved highest F1 (0.863) but exhibited severe vulnerability with 59\% MCC degradation and net impact of -73 (Table~\ref{tab:triplet_effect}). LLM judges showed MCC drops of 50-68\%, while \textbf{embedding methods} demonstrated superior robustness (9-13\% degradation). NLI's minimal disagreements (15, 2\%) belie a 34\% MCC drop due to extreme prediction imbalance (98\% `grounded'), where MCC's sensitivity to true negatives exposes minority class detection failures (TN: 13$\rightarrow$4). Statistical tests confirm significant differences (Cochran's Q: p<0.001; McNemar: 29/55 vs 9/55 pairs), indicating increased variance under noise.

\paragraph{Failure Mechanisms: Over-Reliance on Structured Signals}
Manual inspection revealed systematic error patterns. Triplets helped when correctly extracting semantic relationships, making implicit information explicit. Failure modes include: (1) \textbf{temporal misalignment} (most common); (2) \textbf{hallucinated entities}; (3) \textbf{wrong numeric values and cross-context contamination}. LLM-based judges exhibit \textbf{preferential trust of structured signals over textual evidence}, anchoring on KG format even when contradicting prose--a structural bias that amplifies extraction errors. These results reveal that \textbf{success requires both robust evaluation methods and high-quality triplet extraction}. While accurate triplets improve detection by making implicit relationships explicit, current methods lack robustness to noisy signals, leading to systematic degradation. Since these effects emerged from only 5-7 triplets per page, real-world deployments with dozens of triplets will magnify both benefits and vulnerabilities, necessitating detection methods with built-in triplet verification and improved KG extraction pipelines.

\begin{table}[ht]
\centering
\caption{Statistical Significance Tests. Cochran's Q confirms significant differences. McNemar with Holm-Bonferroni correction for pairwise comparisons. Bootstrap CIs: 10,000 resamples.}
\label{tab:statistical_tests}
\small
\begin{tabular}{l r r r r}
\toprule
\textbf{Condition} & \textbf{Cochran's Q} & \textbf{p-value} & \textbf{McNemar Sig. Pairs} & \textbf{Total Pairs} \\
\midrule
W/O Triplets & 132.63 & <0.001 & 29 & 55 \\
W/ Triplets & 73.85 & <0.001 & 9 & 55 \\
\bottomrule
\end{tabular}
\end{table}

\section{Discussion \& Limitations}

\begin{itemize}
  \item \textbf{Annotation limitations:} Validators were in-house AI engineers; the task was designed to minimize finance domain expertise requirements by relying on explicit evidence linkage, but it also acknowledges that some subtle factual judgments may require domain expertise.
  \item \textbf{Task complexity:} Our benchmark focuses on extractive QA tasks to enable controlled evaluation of detection methods. Real-world financial analysis involves multi-hop reasoning, complex numerical calculations, and cross-document synthesis-extensions that would strengthen future versions of this benchmark.
  \item \textbf{KG extraction quality:} We applied FinReflectKG's automated validation mechanisms loosely rather than strictly to retain naturally occurring extraction errors for robustness testing (see Introduction). Our manual inspection identified four systematic error patterns (temporal misalignment, hallucinated entities, wrong values, cross-context contamination), though comprehensive characterization of all error types across the dataset remains future work.
  \item \textbf{Scale:} Our benchmark's scale (n=755) represents the final annotated dataset after filtering from $>$5{,}000 initially generated examples. Despite this modest size, statistical power is sufficient for the reported comparisons (bootstrap confidence intervals and pairwise significance tests confirm robust effect sizes).
\end{itemize}

\section{Conclusion \& Future Work}

This work introduced FinBench-QA-Hallucination, a benchmark for evaluating hallucination detection in knowledge-graph-augmented financial question answering. Our evaluation of six detection approaches on 755 annotated SEC 10-K examples reveals that while LLM-based judges and embedding methods achieved strong performance in clean conditions (Table~\ref{tab:best_scores}), noisy KG signals substantially degraded most detectors' reliability (Table~\ref{tab:statistical_tests}), with embedding-based methods demonstrating the greatest robustness.

\paragraph{Future Directions:}
Key extensions include: (1) multi-hop reasoning tasks requiring synthesis across document sections, temporal reasoning, and numerical calculations; (2) claim-level verification with tool-augmented detection (calculators, unit converters, entity resolution); (3) expert annotation to capture domain-specific reasoning failures; and (4) systematic KG quality quantification to enable targeted robustness evaluations.

By providing a controlled, reproducible benchmark with explicit KG noise conditions, this work establishes a foundation for developing more robust hallucination detection methods suitable for high-stakes financial applications where factual accuracy is paramount.

\appendix

\section{Input Format Specifications}

To ensure reproducibility and clarity, we provide the exact input formats used for each detection approach.

\subsection{All Approaches}

\textbf{Input format (WITHOUT triplets):}
\begin{verbatim}
# Context
[Raw context text from SEC 10-K page]
\end{verbatim}

\textbf{Input format (WITH triplets):}
\begin{verbatim}
# Context
[Raw context text from SEC 10-K page]

# Triplets
Triplet 1: ["Triplet 23", ["PG", "ORG", "Guides_On",
            "4%-FY-2024", "MERGED_NUMERIC_UNIT_PERIOD"]]
Triplet 2: ["Triplet 16", ["Fabric & Home Care", "SEGMENT",
            "Has_Value_Merged", "1%-NA-FY-2024", ...]]
...
\end{verbatim}

\section{Question-Answer Generation Prompt}
\label{sec:qa-gen-prompt}

The complete prompt used for generating question-answer pairs with Qwen-3-235B is provided below verbatim:

\begin{small}
\begin{verbatim}
You are an expert at creating meaningful, verifiable questions from
a given text context.

# Input you will receive:
- context (text chunk). It is a Markdown text which can have markdown
  tables in it. You have to very precise in reading this context and
  understanding it.
- a list of triplets where each is a list of values. These triplets
  are made from this context only BUT there can be issues. Any Triplet
  contradicting the information from context text, don't use it to
  generate question.

# Task:
- Produce multiple high-quality questions that are answerable **only**
  and only by using:
  1) one or more of the exact triplets and
  2) context line snippets (returned as the snippet's exact starting
     words, 2-5 words if multi word lines).

Hard constraints (must follow exactly):
1) Use ONLY the provided context and the triplets. No external
   knowledge or assumptions. Triplets are from text only and try to
   look at triplets and context both to create meaningful questions.
2) Each question and correct_answer must be directly supported by
   the triplet(s) and the cited snippet(s) from context. No inference,
   no computation, no filling-in. Do not use "according to triplet",
   "according to context etc", "as given in text" etc etc. Ask a good
   quality question. Use Ticker company info to avoid generic and
   ambiguous questions which may be aplied to multiple companies or
   years at once
3) The snippets MUST be returned as its exact starting words. I say
   it again, ONLY the "EXACT" "STARTING" "FEW" WORDS and not the
   exact whole sentences. Don't say "Word 1 Word2 ....". Just EXACT
   words. For Example: ["Revenue of year", " | $20,434 | 2020",
   "$(53232)"] etc. If the answer contains any numbers, definitely
   include all those numbers for sure in the snippets
4) The correct_answer must must be from the cited snippet AND MUSt
   be SUPPORTED by the Triplet TOO. Do not invent words. Do not use
   external knowledge. Even though Triplets are from the context and
   you can use their info to create questions but the correct_answer
   must come the source text ad must be human readable friendly. You
   CAN NOT give the answer as a weird Triplet entity such as:
   '3.5bn-EUR-Q1-FY-2024'
5) For each question, specify which exact list of triplet(s) were
   used in `triplets_used`.
6) Avoid duplicate or paraphrased questions. Each question should be
   unique and meaningful. Pay close attention to any triplet having
   numbers or names. Those are very important snd meningful question
   can be made from them
7) If you cannot produce any valid or informative questions under
   these constraints, return an empty json

Hints for creating good quality questions:
1. Triplets are extracted from the same context text you have been
   given nd are in format: (Subject OR Head Entity, Head Type,
   Relationship, Object OR Tail Entity, Tail Type)
2. Pay close attention to any triplet having numbers, names and more
   specifically having MERGED_NUMERIC_UNIT_PERIOD. Those are meningful
   question can be made from them. Other type of triplets can be good
   too so be mindful to refer to Triplet AND the Text both at the
   same time.
3. MERGED_NUMERIC_UNIT_PERIOD = "A unified representation of a
   financial metric, combining value, unit, and period into a single
   string. The format is VALUE-UNIT[-PERIOD]. Use 'mn' or 'bn' only
   when explicitly indicated in the source (cell value,
   caption/header/footnote); never use 'k'. Otherwise store absolute
   values with '-USD', '-shares', or '-count'. Examples:
   2000000-USD-NA, 2mn-USD-FY-2023, 3.5bn-EUR-Q1-FY-2024, 50-bps-NA,
   21.5%-FY-2022, 125bps-2023, 25mn-shares-Q4-FY-2023, 3-count-NA."

Example:
A triplet looking like: ["Capital expenditures", "FIN_METRIC",
"Has_Value_Merged","0.7bn-USD-FY-2024","MERGED_NUMERIC_UNIT_PERIOD"]
means that the Company (ticker's) Capital Expenditure in the Financial
year 2024 was 0.7 billion US Dollar. You can easily make meaningful
questions from it.

Important NOTE:
There can be chances that a triplet might be wrong so create questions
and answers for the one where the Triplet is fully supported by the
context else it'll be a wrong question where Triplet is contradicting
the Text

Now read the context and triplets carefully and after deep thinking,
produce the response object.
\end{verbatim}
\end{small}

\section{Question-Answer Evaluation Prompt}
\label{sec:qa-eval-prompt}

The complete prompt used for evaluating the quality of generated question-answer pairs (Keep/Drop decisions) is provided below verbatim:

\begin{small}
\begin{verbatim}
You are a quality evaluator for financial Q&A pairs generated from
SEC-10K filing triplets.

Your job is to evaluate each generated Q&A pair and decide: **Keep**
or **Drop**.

## Evaluation Criteria

**DROP if ANY of these apply:**

1. **Question Issues:**
   - Too generic/broad (e.g., "What metrics did Apple disclose in
     2022?")
   - Missing company name/ticker OR year when context provides them
   - Asks for information not directly available in the triplet
   - Uses phrases like "triplet", "context", "record", "provided data"

2. **Answer Issues:**
   - Contains information not in the triplet
   - Makes assumptions or adds external knowledge
   - Reasoning doesn't clearly cite which triplet fields were used
   - Answer contradicts the triplet data

3. **Overall Quality:**
   - Too simple/obvious (e.g., "When did Company X file?" when every
     company files annually)
   - Not factual enough to be useful
   - Ambiguous across multiple companies/years

**KEEP only and only if:**
- Question is specific and factual
- Directly answerable from triplet alone
- Includes company and year context
- Answer and Answer reasoning clearly maps triplet fields to answer
- Provides useful, non-obvious information

## Output Format

For each Q&A pair, respond with:
```
Decision: Keep/Drop
Reason: [Brief explanation if Drop, "Meets quality standards" if Keep]
```

## Examples

**DROP Example:**
```
Q: What did the company disclose in 2022?
A: Financial metrics
Decision: Drop
Reason: Question too generic, answer too vague, missing company context
```

**KEEP Example:**
```
Q: Who was the CEO of Apple (AAPL) in October 2022?
A: Timothy D. Cook
Decision: Keep
Reason: Meets quality standards
```

Evaluate strictly. When in doubt, DROP.
\end{verbatim}
\end{small}

\section{Company Coverage}
\label{sec:company-coverage}

List of of 57 S\&P 100 companies from the 755 valid examples is provided below:

AAPL, ABBV, ABT, AMD, AMGN, AMZN, BAC, BKNG, BLK, BMY, CAT, CL, COP, COST, CSCO, CVX, DHR, DIS, DOW, DUK, EMR, F, GD, GE, GILD, GM, HON, IBM, INTC, JNJ, JPM, LLY, LMT, MA, MDT, MO, MS, MSFT, NEE, NFLX, NKE, PEP, PFE, PG, PM, PYPL, QCOM, RTX, SBUX, SCHW, T, TSLA, TXN, UNH, UNP, VZ, XOM.

\begin{figure}
    \centering
    \includegraphics[width=0.99\linewidth]{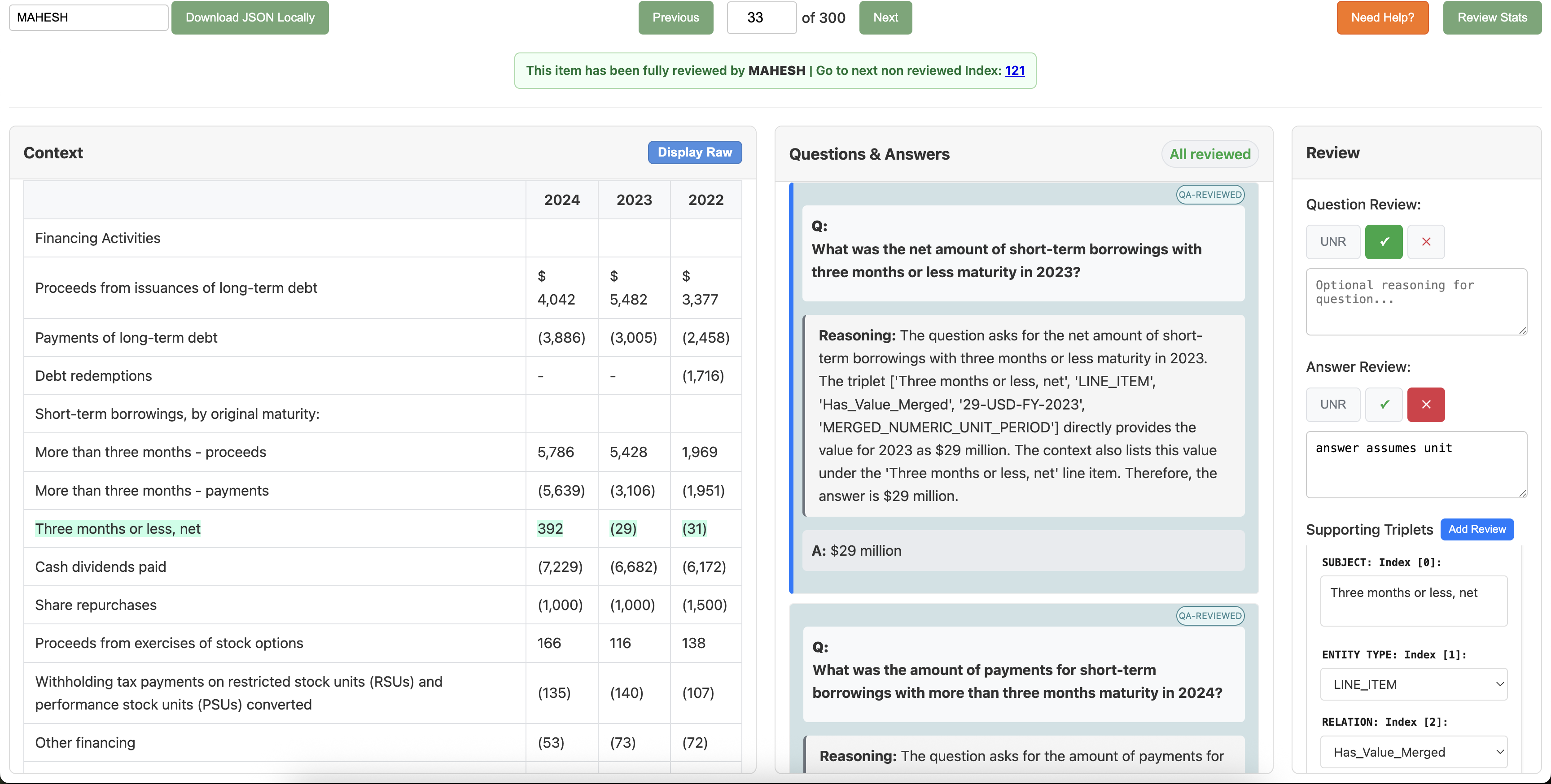}
    \caption{Annotation App Interface}
    \label{fig:annotation_interface}
\end{figure}

\clearpage
\bibliographystyle{unsrt}
\bibliography{references}

\end{document}